\pdfoutput=1

\documentclass[11pt]{article}

\usepackage[preprint]{acl}

\usepackage{times}
\usepackage{latexsym}

\usepackage[T1]{fontenc}

\usepackage[utf8]{inputenc}

\usepackage{microtype}

\usepackage{inconsolata}

\usepackage{graphicx}

\usepackage{booktabs}
\usepackage{xcolor}
\usepackage{colortbl}
\usepackage{xspace}
\usepackage{amsmath}
\usepackage{graphicx}
\usepackage{subcaption} 

\usepackage[capitalize]{cleveref}
\crefname{section}{Sec.}{Secs.}
\Crefname{section}{Section}{Sections}
\Crefname{table}{Table}{Tables}
\crefname{table}{Tab.}{Tabs.}

\usepackage{enumitem} 
\setlist[enumerate]{itemsep=0.5em, topsep=0.5em, parsep=0em, partopsep=0em}

\newcommand{\bench}{DynamicBench\xspace}

\title{DynamicBench: Evaluating Real-Time Report Generation \\in Large Language Models}

\author{
  \textbf{Jingyao Li\textsuperscript{1}},
  \textbf{Hao Sun\textsuperscript{2}},
  \textbf{Zile Qiao\textsuperscript{2}}\thanks{Corresponding Author},
  \textbf{Yong Jiang\textsuperscript{2}},
  \\
  \textbf{Pengjun Xie\textsuperscript{2}},
  \textbf{Fei Huang\textsuperscript{2}},
  \textbf{Hong Xu\textsuperscript{1}},
  \textbf{Jiaya Jia\textsuperscript{3}}
  \vspace{2mm}
\\
  \textsuperscript{1}The Chinese University of Hong Kong,
  \textsuperscript{2}Alibaba Group,
\\
  \textsuperscript{3}The Hong Kong University of Science and Technology
}

\begin{document}
\maketitle


\begin{abstract}
Traditional benchmarks for large language models (LLMs) typically rely on static evaluations through storytelling or opinion expression, which fail to capture the dynamic requirements of real-time information processing in contemporary applications. To address this limitation, we present \bench, a benchmark designed to evaluate the proficiency of LLMs in storing and processing up-to-the-minute data. \bench utilizes a dual-path retrieval pipeline,  integrating web searches with local report databases. It necessitates domain-specific knowledge, ensuring accurate responses report generation within specialized fields. By evaluating models in scenarios that either provide or withhold external documents, \bench effectively measures their capability to independently process recent information or leverage contextual enhancements. Additionally, we introduce an advanced report generation system adept at managing dynamic information synthesis. Our experimental results confirm the efficacy of our approach, with our method achieving state-of-the-art performance, surpassing GPT4o in document-free and document-assisted scenarios by 7.0\% and 5.8\%, respectively. The code and data will be made publicly available.
\end{abstract}

\begin{figure*}[t]
    \centering
    \includegraphics[trim={25 120 150 150}, clip, width=\linewidth]{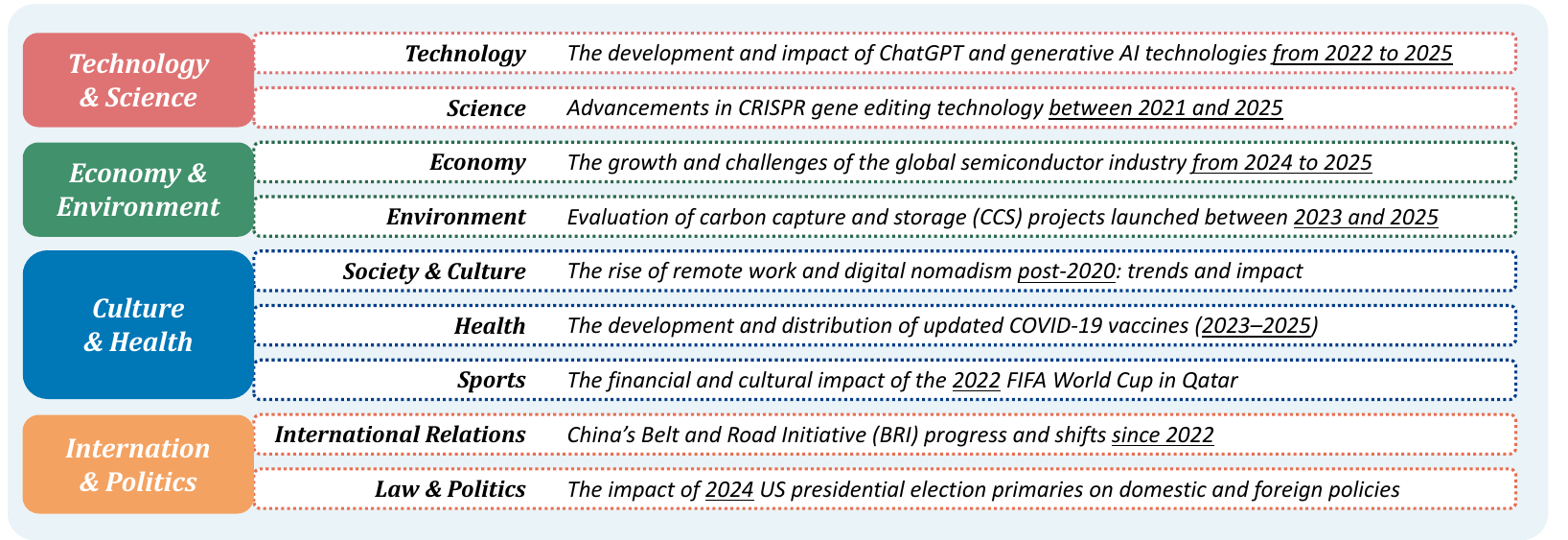}
    \caption{Query examples across four major categories: \textit{Tech \& Science}, \textit{Economy \& Environment}, \textit{Culture \& Health}, and \textit{International \& Politics}, each with multiple subcategories. }
    \label{fig:query-examples}
\end{figure*}

\begin{figure}[t]
    \centering
    \includegraphics[trim={70 160 600 170}, clip, width=\linewidth]{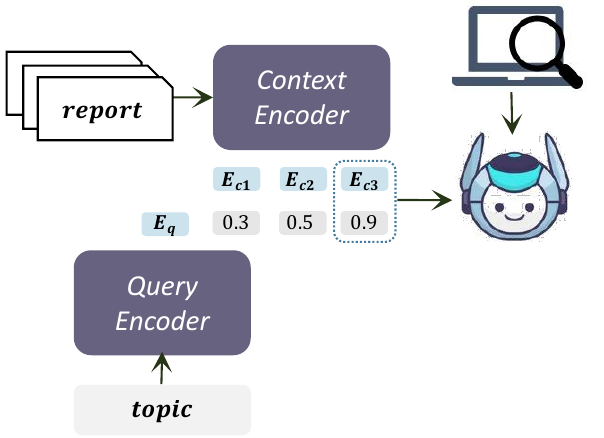}
    \caption{Dual-Path Retrieval Report Generation System that combines retrieval-augmented generation (RAG) from a local financial report database and Web Search to gather information. The related information are fed into a LLM for comprehensive report generation.}
    \label{fig:rag}
\end{figure}

\begin{figure*}[t]
    \centering
    \includegraphics[trim={190 230 190 120}, clip, width=\linewidth]{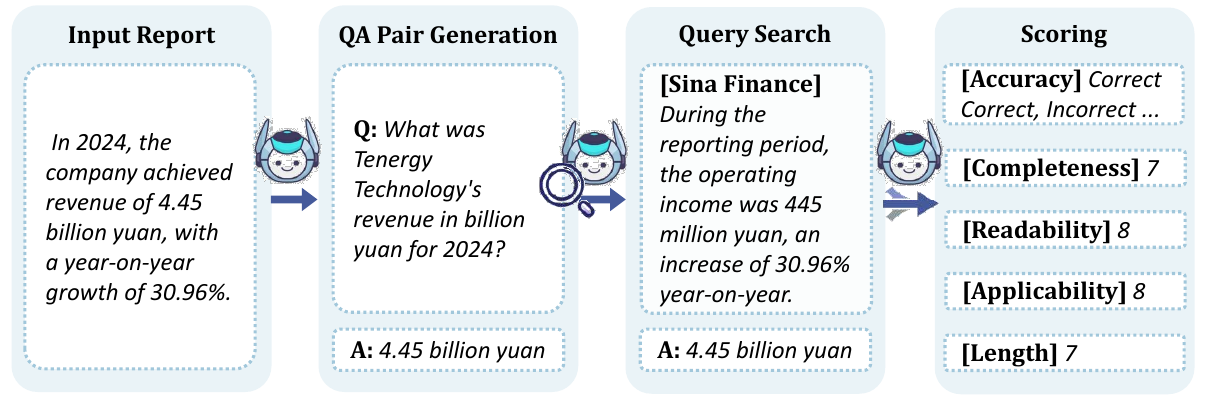}
    \caption{The evaluation system process begins with the generation of question and answer (Q\&A) pairs from key details extracted from the input report, which are used as queries in a dual-path retrieval strategy. This strategy involves searching both online and within a local financial report database to gather relevant information. The system assesses the accuracy of each Q\&A pair by aligning reported data with retrieved information and calculates the \textit{accuracy}. The \textit{completeness}, \textit{readability}, \textit{applicability}, and \textit{length} of the report are also evaluated based on the retrieved information.}
    \label{fig:eval-framework}
\end{figure*}

\begin{figure*}[t]
    \centering
    \includegraphics[trim={25 40 150 155}, clip, width=\linewidth]{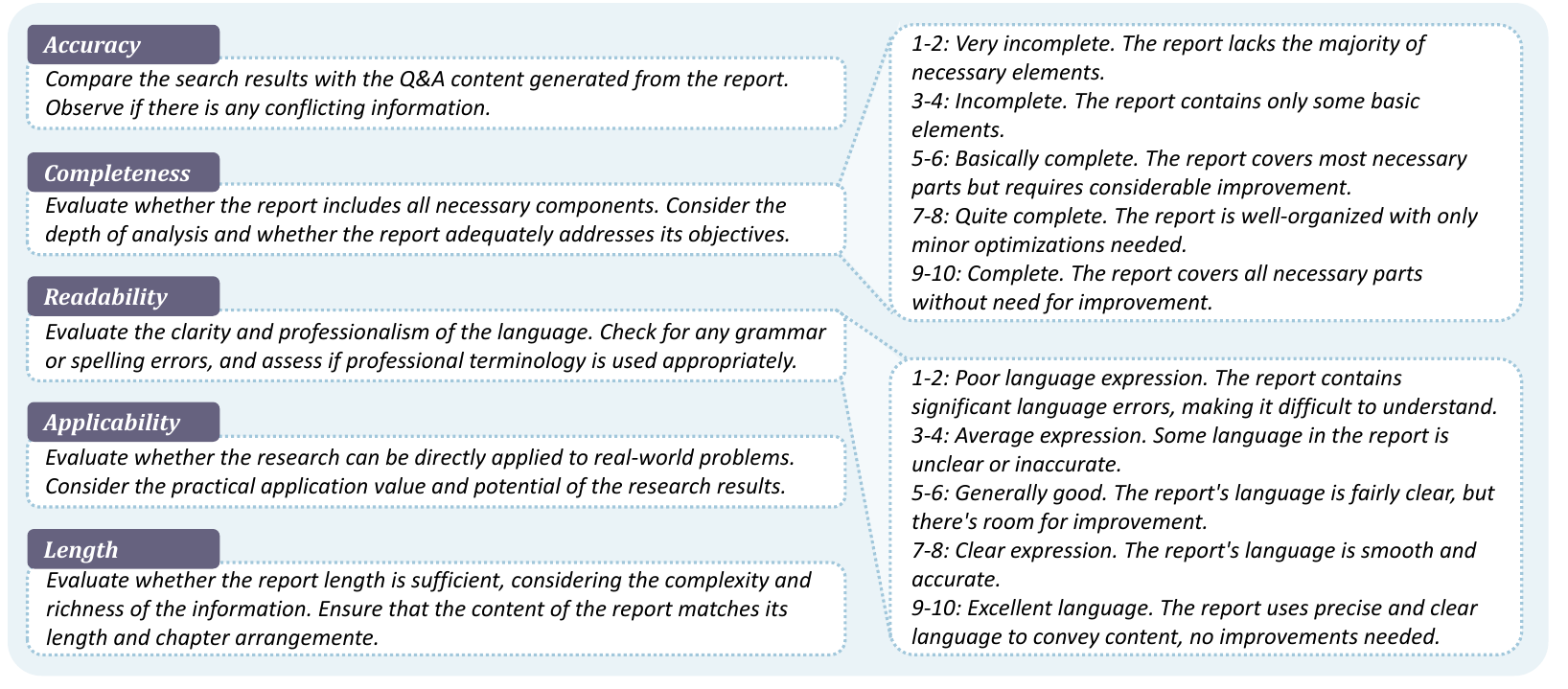}
    \caption{Evaluation criteria for report analysis, focusing on five distinct metrics: \textit{accuracy}, \textit{completeness}, \textit{readability}, \textit{applicability}, and \textit{length}. Each criterion is supported by a detailed description to guide evaluators in comparing search results, examining the depth of analysis, assessing linguistic clarity and professionalism, evaluating real-world applicability, and verifying the adequacy of report length relative to content richness. The right panels examplify rating scale from 1 to 10 for \textit{completeness} and \textit{readability}, offering specific guidelines on how each score reflects the report's quality and coherence.}
    \label{fig:eval-levels}
\end{figure*}

\begin{figure*}[t]
    \centering
    \includegraphics[trim={190 230 190 120}, clip, width=\linewidth]{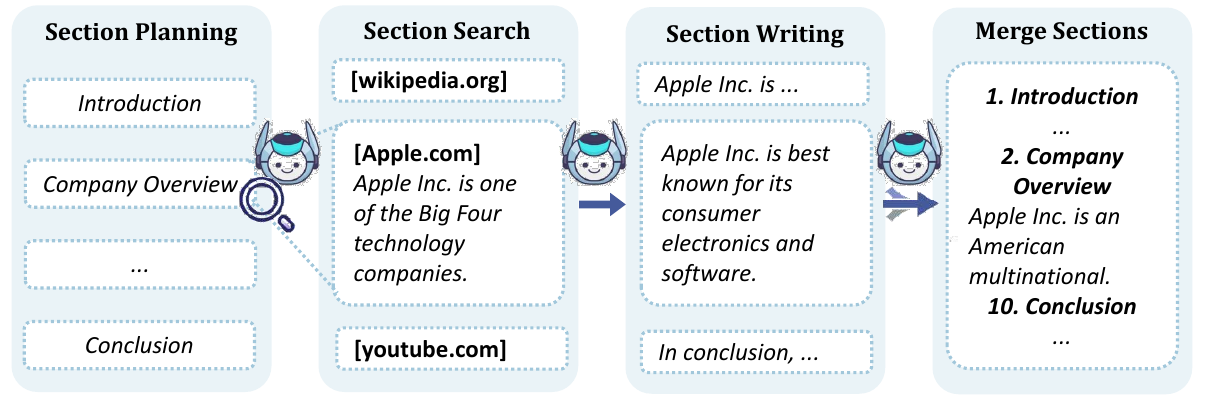}
    \caption{Report composition workflow of topic \textit{Financial performance review of Apple Inc. in 2021}: The four-step process begins with \textit{section planning}, where major section titles are established based on the research topic. This is followed by \textit{section search}, which involves precise queries to gather relevant evidence from local and online sources for each section. \textit{Section writing} utilizes the collected evidence to create detailed analysis texts, complete with tables and charts. Finally, \textit{merge sections} compiles all section analyses into a cohesive report, optimizing narrative flow and outputting a complete document.}
    \label{fig:report-framework}
\end{figure*}

\section{Introduction}

In recent years, Large Language Models (LLMs) have revolutionized natural language processing, displaying exceptional proficiency in tasks ranging from language generation to contextual comprehension across various domains. However, traditional benchmarks remain confined to static evaluations, often relying on storytelling or expression of opinion. Such static, subjective assessment criteria fail to capture the dynamic nature of real-time information processing, which is crucial for understanding the true capabilities of LLMs~\cite{writingbench, hellobench}.

Addressing these limitations, we introduce \bench, a benchmark designed to evaluate LLMs' proficiency in acquiring and processing real-time data. Distinguished by its demand for contemporary information retrieved through web searches and database queries, \bench necessitates that models possess the most up-to-date knowledge for accurate responses. Utilizing a dual-path retrieval pipeline, \bench combines local report databases with web searches, ensuring access to comprehensive data for thorough report evaluation. \bench assesses a wide array of domains, capturing the latest dynamics across critical categories such as \textit{Tech \& Science}, \textit{Economy \& Environment}, \textit{Culture \& Health}, and \textit{International \& Politics}. Through both scenarios, providing or withholding external documents, \bench evaluates a model's capability to store knowledge or process recent external information effectively. This requirement for precise data collection within specialized fields guarantees the accuracy and objectivity of the evaluation process, bridging the gap in current methodologies regarding objective and real-time assessments.

Beyond the benchmark itself, our contribution includes a robust solution for report generation, adept at tackling the complex challenges posed by dynamic information generation. Our system begins with report planning based on the query followed by query generation and resource aggregation using a dual-path retrieval pipeline from both local and online data. The system self-assesses whether further information gathering is necessary and ensures adequate information collection, informing detailed report writing that integrates tables and charts for enhanced clarity. Ultimately, it outputs a comprehensive, coherent report that reflects the latest data. Experimental results demonstrate the efficacy of our methods. We evaluate LLMs under two conditions: without and with document assistance, and analyze their performance across different domains in both scenarios. Our approach showcases state-of-the-art performance across several metrics, surpassing GPT4o by 7.0\% and 5.8\%, respectively.

In summary, our contributions are as follows:
\begin{enumerate}
    \item We introduce \bench, a novel benchmark that evaluates LLMs based on real-time information acquisition and processing capabilities, utilizing a dual-path retrieval system that combines local and online data sources.
    \item We develop a comprehensive report generation system that plans, searches, and writes detailed reports, ensuring the integration of up-to-date information for accurate and coherent documentation.
    \item We demonstrate through experimental results the advanced capabilities of our approach, which achieves state-of-the-art performance compared to leading LLMs, highlighting significant improvements across multiple metrics.
\end{enumerate}

\section{Related Works}
\subsection{Writing Benchmarks}
Recent advancements in evaluating Large Language Models (LLMs) have led to the creation of several benchmarks aimed at assessing different aspects of language generation and comprehension. LongBench-Write~\cite{longwriter} focuses on understanding model capabilities in adhering to complex writing tasks within LLMs. HelloBench~\cite{hellobench} expands evaluation efforts by categorizing long text generation into distinct tasks such as open-ended QA and heuristic text generation. EQ-Bench~\cite{eqbench} introduces an evaluation of emotional intelligence by assessing LLMs' abilities to comprehend and predict emotional intensities in dialogues. WritingBench~\cite{writingbench} offers a comprehensive evaluation across domains and subdomains, including creative and technical writing. These traditional methods which predominantly focused on storytelling or opinion expression, adopting static and subjective evaluation criteria. In comparison, our system not only offers a holistic framework that covers a wide range of topics and evaluates various aspects of writing, but also utilizes real-time web searches and database queries to access the latest information. Thus, our system evaluates models' ability to process and utilize real-time information effectively. Moreover, our benchmark necessitates constructing precise reports within specialized fields, thus ensuring the accuracy and objectivity of the information utilized. These attributes enable our benchmark to bridge the gap in the current benchmarks concerning objective and real-time assessments.

\subsection{Long-Context Capabilities of LLMs}
Large Language Models (LLMs) such as Claude-3~\cite{claude3}, DeepSeek-R1~\cite{deepseekr1}, DeepSeek-v3~\cite{deepseekv3}, GPT-4o~\cite{gpt4}, and Qwen-2.5~\cite{qwen25} have demonstrated remarkable capabilities in various domains, including understanding and generating complex language tasks. These models serve as foundational tools for numerous applications, yet often face limitations in generating extended outputs or adhering to intricate task constraints. LongWriter~\cite{longwriter} addresses the output length limitation in current LLMs by proposing AgentWrite, an agent-based pipeline that enables models to generate coherent outputs exceeding 20,000 words. Suri~\cite{suri} introduces a multi-constraint instruction-following approach for generating long-form texts. It can generate significantly longer texts with sustained quality and compliance to constraints. In contrast, our work surpasses previous efforts by effectively generating extended content with enhanced coherence and quality. 

\section{Methodology}
In order to address the challenges posed by the dynamic information generation and the need for accurate report construction, our methodology centers around the development of a benchmark and a robust system solution. In \cref{sec:benchmark}, we introduce a benchmark is designed to assess the ability of LLMs in acquiring and processing real-time data. In \cref{sec:report_generation}, we propose our report generation system solution.

\subsection{\bench}
\label{sec:benchmark}
Traditionally, benchmarks~\cite{writingbench, hellobench} have relied on storytelling or opinion expression, which are non-time-sensitive due to their static nature. In contrast, our benchmark, as exemplified in ~\cref{fig:query-examples}, requires contemporary, time-sensitive information retrieved via web search and database queries. This approach necessitates the possession of the most up-to-date domain-specific knowledge for accurate responses, thus assessing the capability of current models in acquiring and processing real-time information. Moreover, unlike traditional subjective evaluations, our benchmark demands the collection of data to construct reports within specialized fields, ensuring the accuracy and objectivity of the evaluation utilized. These attributes position our benchmark to narrow the gap in the current benchmarks regarding objective and real-time assessments. 
Our benchmark comprises the following categories:

\begin{itemize}
    \item \textbf{Tech \& Science}: \textit{technology} and \textit{science}.
    \item \textbf{Economy \& Environment}: \textit{economy} and \textit{environment}.
    \item \textbf{Culture \& Health}: \textit{society and culture}, \textit{health}, and \textit{sports}.
    \item \textbf{International \& Politics}: \textit{international relations} and \textit{law and politics}.
\end{itemize}

\subsubsection{Information Retrieval Process}
To construct our local report database, we sourced 148,589 annual reports from 10,338 global companies from AnnualReport\footnote{\url{https://www.annualreports.com}}. These reports cover a diverse array of domains including economy, environment, technology, science, culture, health, laws, politics, etc. We leverage a retrieval-augmented generation (RAG) approach to perform information retrieval, as illustrated in~\cref{fig:rag}.

The process involves using a context encoder to encode the local report database and a query encoder to encode incoming queries. Each report and query is transformed into embeddings, denoted as $\mathbf{E}_c$ for context and $\mathbf{E}_q$ for queries. The similarity between these embeddings is computed using cosine similarity, defined as:

\begin{equation}
    \text{Similarity}(\mathbf{E}_c, \mathbf{E}_q) = \frac{\mathbf{E}_c \cdot \mathbf{E}_q}{\|\mathbf{E}_c\| \|\mathbf{E}_q\|}
\end{equation}

The system effectively extracts the report block with the highest similarity score as the most relevant information. This process is mathematically represented as selecting the block $\mathbf{B}^*$ such that:

\begin{equation}
\mathbf{B}^* = \operatorname{argmax}_i \, \text{Similarity}(\mathbf{E}_{c_i}, \mathbf{E}_q)
\end{equation}

We utilize a dual-path retrieval pipeline, obtaining information through both the local report database and web searches. This comprehensive approach ensures that our system leverages the available data for robust and comprehensive report generation.

\subsubsection{Evaluation Process}

As illustrated in \cref{fig:eval-framework}. The initial stage of our evaluation process involves extracting key information from the input report to generate question-and-answer (Q\&A) pairs. These pairs form the basis for subsequent information retrieval, wherein queries derived from these pairs are employed within a dual-path retrieval strategy. This strategy utilizes both web searches and the local financial report database to gather comprehensive information. Once retrieved, our system evaluates several metrics, as depicted in \cref{fig:eval-levels}. These metrics include:

\paragraph{Accuracy.} The accuracy of each Q\&A pair is determined by assessing the alignment between reported data and the information retrieved; If the search data corroborates the Q\&A or no discrepancies are found, the system labels it as \textit{Correct}. Conversely, if relevant content is missing or conflicts are detected, it may be marked as \textit{Cannot Determine} or \textit{Incorrect}. The average accuracy across all queries is calculated to determine the final accuracy metric of the report. 

\paragraph{Completeness.} This metric assesses whether the report includes all necessary elements and adequately addresses its objectives. Evaluators use a rating scale to determine completeness, from very incomplete (1-2 points) to fully complete (9-10 points).

\paragraph{Readability.} This criterion evaluates the clarity and professionalism of the report's language, checking for grammatical and spelling errors, as well as the appropriate use of professional terminology. Readability is rated from poor language expression (1-2 points) to excellent language use (9-10 points).

\paragraph{Applicability.} This metric gauges the practical application value of the research findings, assessing whether the report can directly contribute to solving real-world problems. Applicability is ranked from poor applicability (1-2 points) to significant application value (9-10 points).

\paragraph{Length.} This criterion evaluates if the report's length sufficiently covers the complexity and richness of the information presented. Length is rated from highly insufficient (1-2 points) to perfectly sufficient (9-10 points), considering the adequacy of each chapter's content.

\subsection{Report Generation Process}
\label{sec:report_generation}
In addressing complex report generation challenges, our system provides a structured methodology for high-quality output, as depicted in \cref{fig:report-framework}.

\paragraph{Section Planning.} This initial phase involves the establishment of major section titles based on the research topic. For instance, in reviewing Apple Inc.'s financial performance in 2021, sections such as Introduction, Company Overview, and Conclusion are identified to organize the report logically.

\paragraph{Section Search.} For each section, the model initially generates $K$ queries aimed at retrieving relevant data and insights. These queries are used to search both local databases and online resources, aggregating the retrieved content. The model then conducts a self-assessment of the gathered information to determine if it is sufficient for drafting the section. If the content is deemed insufficient, additional queries are generated and executed to fill any gaps in information. This iterative process continues until ample data is acquired, allowing the system to proceed to the next stage.

\paragraph{Section Writing.} Utilizing the collected evidence, the system generates detailed analysis texts for each section. This phase includes the integration of tables and charts, enhancing the report's informative quality and visual clarity. 

\paragraph{Merge Sections.} The final step involves compiling all developed sections into a cohesive report. The system optimizes narrative flow, ensuring that the document presents a comprehensive, coherent analysis of the research topic, concluding with a finalized output ready for dissemination.

This systematic approach guarantees thorough, data-driven report creation to produce high-quality documents.

\begin{table*}[t]
\setlength{\tabcolsep}{2.5mm}
\begin{tabular}{lrrrrr|r}
\toprule
Models & Acc. & Comp. & Read. & App. & Len. & Average \\
\midrule
\textbf{w/o Doc} \\
\midrule
DeepSeek-R1~\cite{deepseekr1} & 40.8 & 62.0 & 77.6 & 69.3 & 52.3 & 60.4 \\
DeepSeek-v3~\cite{deepseekv3} & 44.1 & \underline{65.9} & 77.4 & 69.9 & 64.6 & 64.4 \\
Qwen2.5-72B-Instruct~\cite{qwen25} & 49.3 & 61.3 & 75.8 & 68.1 & 60.8 & 63.1 \\
GPT4o~\cite{gpt4} & 58.2 & 65.7 & 77.3 & 70.4 & \underline{66.0} & \underline{67.5} \\
Claude3.7-Sonnet~\cite{claude3} & 55.0 & 64.7 & \textbf{79.0} & \underline{71.3} & 64.5 & 66.9 \\
Suri~\cite{suri} & 43.9 & 45.5 & 62.6 & 63.0 & 43.0 & 51.6 \\
LongWriter~\cite{longwriter} & \underline{68.0} & 45.4 & 62.4 & 62.8 & 41.5 & 56.0 \\
\midrule
\textbf{with Doc} \\
\midrule
DeepSeek-R1~\cite{deepseekr1} & 15.3 & 47.0 & 53.0 & 64.4 & 42.9 & 44.5 \\
DeepSeek-v3~\cite{deepseekv3} & 59.9 & \underline{69.4} & 77.2 & 70.1 & \underline{68.3} & 69.0 \\
Qwen2.5-72B-Instruct~\cite{qwen25} & 63.6 & 60.3 & 75.4 & 66.8 & 60.4 & 65.3 \\
GPT4o~\cite{gpt4} & 63.4 & 65.6 & 77.3 & 70.5 & 67.0 & 68.7 \\
Claude3.7-Sonnet~\cite{claude3} & \underline{69.3} & 65.9 & \textbf{78.7} & \underline{71.2} & 66.3 & \underline{70.3} \\
Suri~\cite{suri} & 51.7 & 40.2 & 57.2 & 60.9 & 37.2 & 49.5 \\
LongWriter~\cite{longwriter} & 45.0 & 30.1 & 40.2 & 47.1 & 26.8 & 37.8 \\
\midrule
\rowcolor{teal!15}Ours & \textbf{74.8} & \textbf{73.7} & \underline{78.0} & \textbf{71.7} & \textbf{74.4} & \textbf{74.5} \\
\bottomrule
\end{tabular}
\caption{Evaluation metrics for LLMs encompassing various aspects, including Accuracy (Acc), Completeness (Comp), Readability (Read), Applicability (App), and Length (Len). With Doc and w/o Doc indicate whether the models were provided with relevant documents. The best and second-best results are highlighted using bold and underlined formatting.}
\label{tab:evaluation-metrics}
\end{table*}

\section{Experiments}

\paragraph{Baseline Models.} Baseline LLMs include Claude3.7~\cite{claude3}, DeepSeek-R1~\cite{deepseekr1}, DeepSeek-v3~\cite{deepseekv3}, GPT4o~\cite{gpt4}, and Qwen-72B~\cite{qwen25}. In addition, we have introduced capacity-enhanced models, such as LongWriter~\cite{longwriter} and Suri~\cite{suri}. LongWriter leverages unique methodologies such as the AgentWrite pipeline to facilitate coherent text generation exceeding 20,000 words. Similarly, Suri employs a multi-constraint instruction-following strategy to generate significantly longer texts while ensuring quality and compliance with constraints.

\paragraph{Evaluation.} To assess the capabilities of current language models in processing dynamic and real-time data, we employed our newly developed benchmark, \bench. This benchmark surpasses traditional methodologies by focusing on the acquisition and analysis of time-sensitive information. By utilizing web search and database queries, we challenge models to demonstrate their proficiency in handling up-to-date domain-specific queries. This approach provides a comprehensive evaluation of a model's ability to integrate the latest information dynamically and construct accurate reports across various specialized fields. Evaluation dimensions include \textit{accuracy}, \textit{completeness}, \textit{readability}, \textit{applicability}, and \textit{length}.

\begin{figure*}[p]
    \centering
    \begin{subfigure}{\linewidth}
        \includegraphics[trim={30 0 10 0}, clip, width=\linewidth]{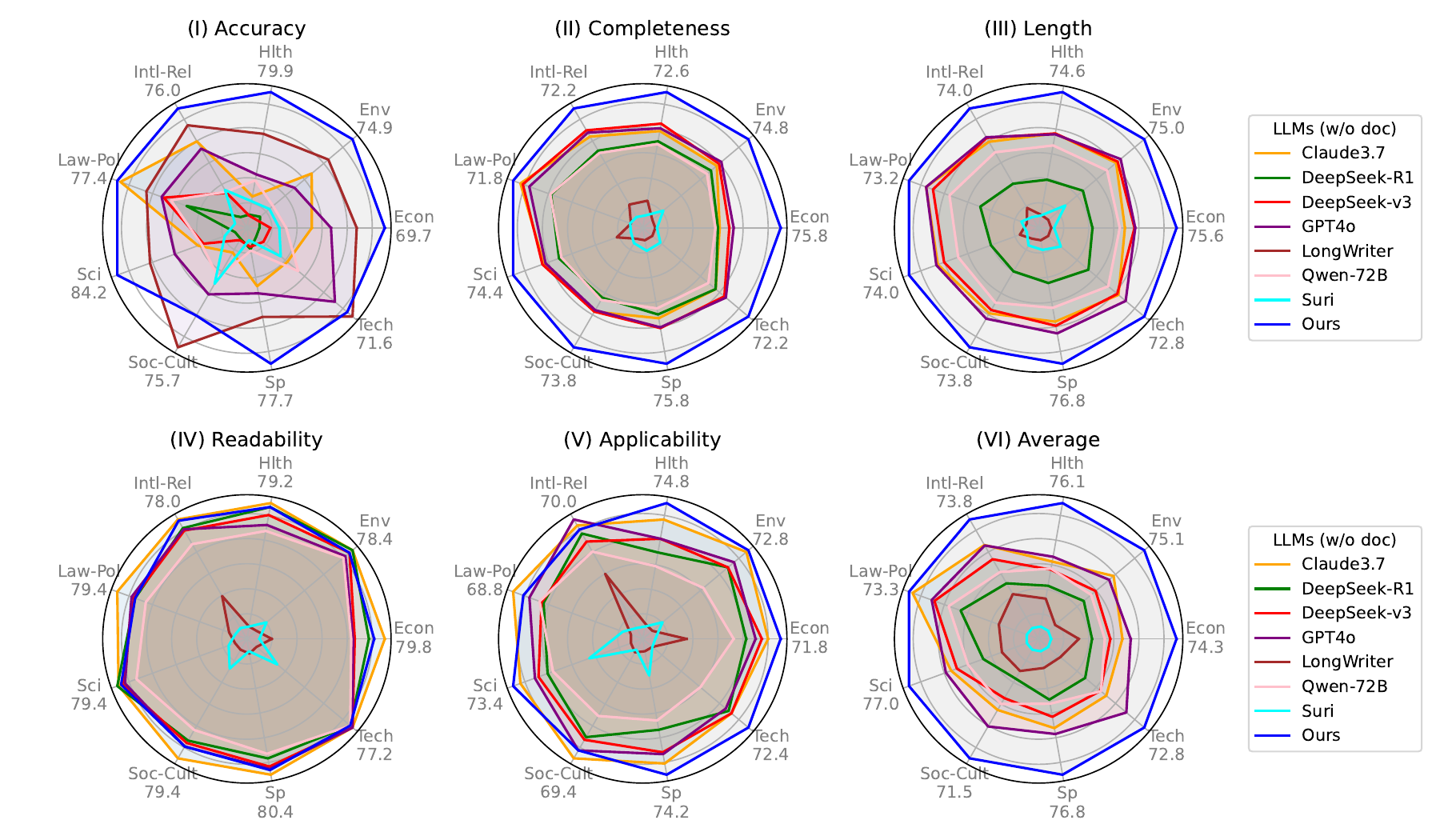}
        \caption{LLMs w/o doc}
        \label{fig:main_result_wo}
    \end{subfigure}

    \begin{subfigure}{\linewidth}
        \includegraphics[trim={30 0 10 0}, clip, width=\linewidth]{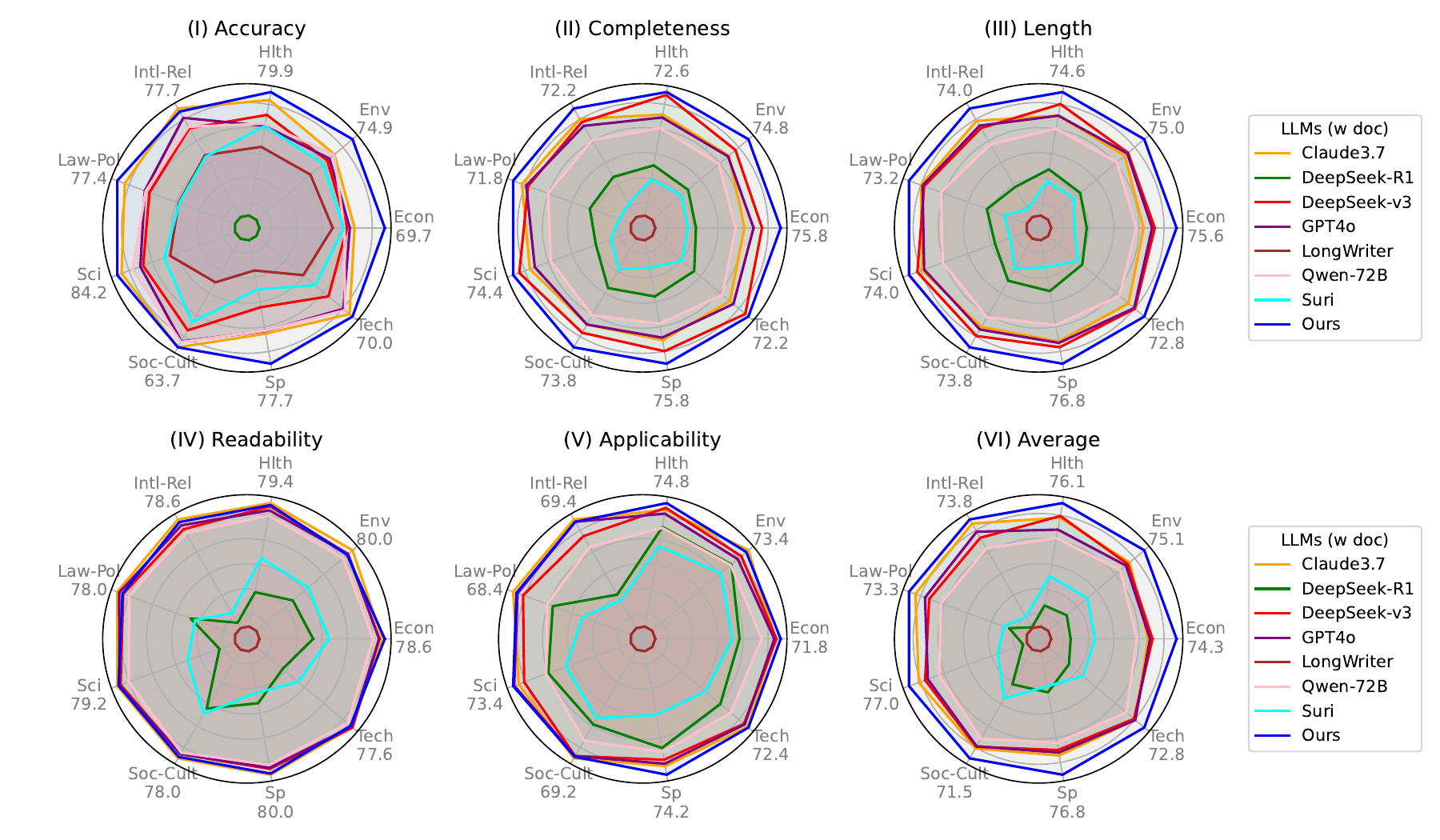}
        \caption{LLMs with doc}
        \label{fig:main_result_w}
    \end{subfigure}
    \caption{Performance of various Systems and LLMs includes Claude3.7~\cite{claude3}, DeepSeek-R1~\cite{claude3}, DeepSeek-v3~\cite{deepseekv3}, GPT4o~\cite{gpt4}, LongWriter~\cite{longwriter}, Qwen-72B~\cite{qwen25}, Suri~\cite{suri}. Evaluation metrics include accuracy, completeness, length, readability, applicability, and average performance. Each model is assessed using a range of topics, such as economy (Econ), environment (Env), health (Hlth), international relations (Intl-Rel), law and politics (Law-Pol), science (Sci), society and culture (Soc-Cult), sports (Sp), and technology (Tech).}
    \label{fig:category-results}
\end{figure*}

\subsection{Results}
In \cref{tab:evaluation-metrics}, we present the outcomes of evaluating our method against baseline models under two conditions: \textbf{w/o doc} and \textbf{with doc}. The \textbf{w/o doc} setting involves baseline LLMs responding without the assistance of external documents, while the \textbf{with doc} setting allows them to utilize our system's dual-path retrieval results. These settings are for assessment of each model's ability to process information independently versus leveraging additional context.

\paragraph{LLMs w/o Doc.} Our method demonstrated new state-of-the-art performance across all dimensions. In terms of \textit{accuracy}, our model achieved 74.8\%, outperforming GPT4o by 16.6\%. The current SOTA in this category was the capability-enhanced model LongWriter, which reached 69.3\%. For \textit{completeness}, our approach attained a score of 73.7\%, exceeding DeepSeek-v3 by 8.1\%. When evaluating \textit{readability}, Claude3.7-Sonnet excelled with a score of 78.7\%, closely followed by our model, which scored 78.0\%. In terms of \textit{applicability}, our model demonstrated a score of 71.7\%, slightly surpassing Claude3.7-Sonnet. Regarding \textit{length}, our model surpassed the competition with a score of 74.4\%, outperforming the next best model, GPT4o, by 7.4\%. Across the five dimensions, our method achieved an average score of 74.5\%, surpassing the current SOTA GPT4o's average by 7.0\%.

\paragraph{LLMs with Doc.} With access to relevant documents, our method continued to showcase state-of-the-art performance across all evaluated metrics. In terms of \textit{accuracy}, our model achieved an impressive 74.8\%, outperforming current SOTA Claude3.7-Sonnet by 5.5\%. For \textit{completeness}, our approach scored 73.7\%, exceeding DeepSeek-v3's score by 4.3\%. Claude3.7-Sonnet led the performance in \textit{readability} with a score of 78.7\%, with our model closely following at 78.0\%. Our model demonstrated superior \textit{applicability}, scoring 71.7\%, which slightly surpassed Claude3.7-Sonnet. In terms of \textit{length}, our model excelled with a score of 74.4\%, significantly outperforming DeepSeek-v3, which scored by 6.1\%. Overall, across the five dimensions, our method achieved an average score of 74.5\%, substantially higher than Claude3.7-Sonnet and GPT4o by 4.2\% and 5.8\%.

\paragraph{Comparison.} The evaluation of models with and without document access reveals notable differences in performance. For general LLMs such as Qwen2.5-72B-Instruct, DeepSeek-v3, GPT4o, and Claude3.7-Sonnet, the performance generally improved significantly when relevant documents were provided. This enhancement highlights LLMs' capacity to leverage external context effectively. Conversely, for capability-enhanced models like Suri and LongWriter, a decline in performance was observed with the inclusion of document inputs. This suggests that these models, which are optimized for generating extended text, may sacrifice some ability to comprehend long contexts when supplied with additional documents. The tendency may result in decreased readability and completeness when external data is introduced. Moreover, both DeepSeek-v3 and Suri, which involve reinforcement learning through human feedback (RLHF) for fine-tuning, exhibited this pattern, indicating that their training methodologies might prioritize generative aspects over contextual understanding.

\subsection{Category-level Analysis}
In \cref{fig:category-results}, we present the detailed results of LLMs and systems across various domains. For \textbf{LLMs w/o doc}, although the average results of previous methods differ significantly from ours, in some domains, better results can be achieved. For example, LongWriter shows slightly higher accuracy in the fields of \textit{Society \& Culture} and \textit{Technology} than ours, and Claude3.7 has slightly better applicability in \textit{Law \& Politics} and \textit{Society \& Culture}. A possible reason for this is that these models develop preferences during training, possibly due to the inclusion of specific knowledge not contained in web searches or the available databases. It is evident that the average results of \textbf{LLMs with doc} show significant improvement compared to \textbf{LLMs w/o doc}, stemming from the supplementary external information enhancing the inherent knowledge of LLMs. While the results of \textbf{LLMs with doc} have narrowed the gap with our system, they generally do not surpass our system, which can be attributed to the fact that both \textbf{LLMs with doc} and our system utilize the same dual-path retrieval information. However, our method effectively leverages this information through a systematic approach.

\section{Conclusion}

This work presents advancements over traditional benchmarks for evaluating large language models (LLMs) by introducing \bench, a dynamic benchmark developed to assess real-time information acquisition and processing capabilities. Utilizing a dual-path retrieval system that synergizes local report databases with web searches, \bench offers comprehensive and objective evaluations across diverse domains. This benchmark demands models to demonstrate domain-specific knowledge, ensuring the generation of accurate reports. Additionally, we have developed an advanced report generation system capable of managing the complexities inherent in dynamic information synthesis. Through systematic planning, query generation, and resource aggregation, this system integrates up-to-date information to produce detailed, coherent reports reflecting the latest data trends. Our experimental results underscore its effectiveness, demonstrating state-of-the-art performance that exceeds existing models like GPT4o across various scenarios.

\clearpage
\section*{Limitations}
While \bench represents an advancement in the evaluation of LLMs by incorporating real-time data retrieval and processing, several limitations remain. Firstly, the dependency on both web searches and local report databases means that the benchmark's effectiveness is contingent on the quality and accessibility of these external sources. Discrepancies or biases in the available data can potentially affect the accuracy and objectivity of the evaluation results. Additionally, the scope of documents considered might not capture the full breadth of contextual knowledge required for specialized fields. The benchmark may not fully assess the depth of understanding necessary for niche domains that require highly specific insights and expertise. These limitations highlight areas for potential improvement, paving the way for future work focused on enhancing data integration strategies, and expanding domain coverage to further advance LLM evaluation and report generation methodologies.

\section*{Broader Impact}
As AI models become increasingly capable of handling real-time information, there are considerations surrounding the ethical use and potential misuse of these technologies. The ability to rapidly generate detailed, coherent reports and real-time data integrations increases the risk of deploying LLMs for misleading or biased content creation. Researchers and developers must prioritize the mitigation of such risks. By fostering transparency and accountability in AI practices, we can ensure that the positive impacts of our work are realized while curtailing the possibilities for harm or misuse. Ultimately, our efforts aim to empower stakeholders with enhanced tools for navigating the complexities of modern information landscapes responsibly.

\section*{AI Assistance Disclosure}
The authors incorporated LLMs to aid in drafting sections of this manuscript. After the initial creation of the text, the authors thoroughly reviewed and refined the material, ensuring its accuracy and integrity, and they assume complete responsibility for the published work.

\bibliography{custom}




\end{document}